\documentclass[10pt,twocolumn,letterpaper]{article}

\usepackage{iccv}
\usepackage{times}
\usepackage{epsfig}
\usepackage{graphicx}
\usepackage{amsmath}
\usepackage{amssymb}
\usepackage{subcaption}
\usepackage{amsfonts}
\usepackage{enumitem}
\usepackage{bbm}
\usepackage{hhline}
\usepackage{multirow}

\usepackage{hhline}
\usepackage{authblk}
\usepackage[breaklinks=true,bookmarks=false]{hyperref}

\iccvfinalcopy 

\ificcvfinal\fi

\begin{document}

\title{

Active Semi-Supervised Learning \\by Exploring Per-Sample Uncertainty and Consistency 
}

\author[1,2]{Jaeseung Lim}
\author[1]{Jongkeun Na}
\author[2]{Nojun Kwak}
\affil[1]{SNUAILAB}
\affil[2]{Seoul National University}

\maketitle
\ificcvfinal\thispagestyle{empty}\fi

\begin{abstract}

  Active Learning (AL) and Semi-supervised Learning are two techniques that have been studied to reduce the high cost of deep learning by using a small amount of labeled data and a large amount of unlabeled data. To improve the accuracy of models at a lower cost, we propose a method called Active Semi-supervised Learning (ASSL), which combines AL and SSL. To maximize the synergy between AL and SSL, we focused on the differences between ASSL and AL. ASSL involves more dynamic model updates than AL due to the use of unlabeled data in the training process, resulting in the temporal instability of the predicted probabilities of the unlabeled data. This makes it difficult to determine the true uncertainty of the unlabeled data in ASSL. To address this, we adopted techniques such as exponential moving average (EMA) and upper confidence bound (UCB) used in reinforcement learning. Additionally, we analyzed the effect of label noise on unsupervised learning by using weak and strong augmentation pairs to address data-inconsistency. By considering both uncertainty and data-inconsistency, we acquired data samples that were used in the proposed ASSL method. Our experiments showed that ASSL achieved about 5.3 times higher computational efficiency than SSL while achieving the same performance, and it outperformed the state-of-the-art AL method.
\end{abstract}

\section{Introduction}

In recent years, deep learning has drastically innovated computer vision and other research areas. However, for a practical deep learning solution, 
a large amount of data is inevitable. For this reason, collecting and labeling a large amount of data has become a major barrier in the development and commercialization of deep learning solutions, requiring significant time and cost. To address this issue, several methods have been researched for decades. Semi-supervised learning (SSL) and active learning (AL) are the two representative methods among them.
\begin{figure}[t]
\begin{center}
\includegraphics[width=0.8\linewidth]{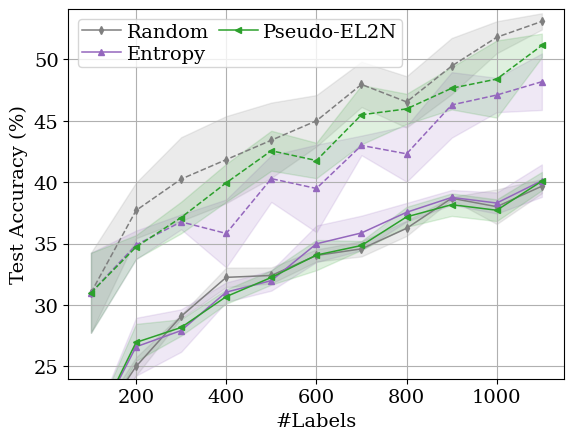}
\end{center}
   \caption{
   Accuracy comparison of random sampling and representative uncertainty-based methods on CIFAR-10. Solid and dashed lines are the results of AL and ASSL, respectively. In the AL case, the difference among the methods is almost negligible. However, in the ASSL case, even though the accuracies are better than those of AL, we can see that the improvement of accuracy for the two uncertainty-based methods is significantly lower than that of random sampling.}
\label{fig:2}
\end{figure}
\begin{figure*}[t]
\begin{center}
\includegraphics[width=0.8\linewidth]{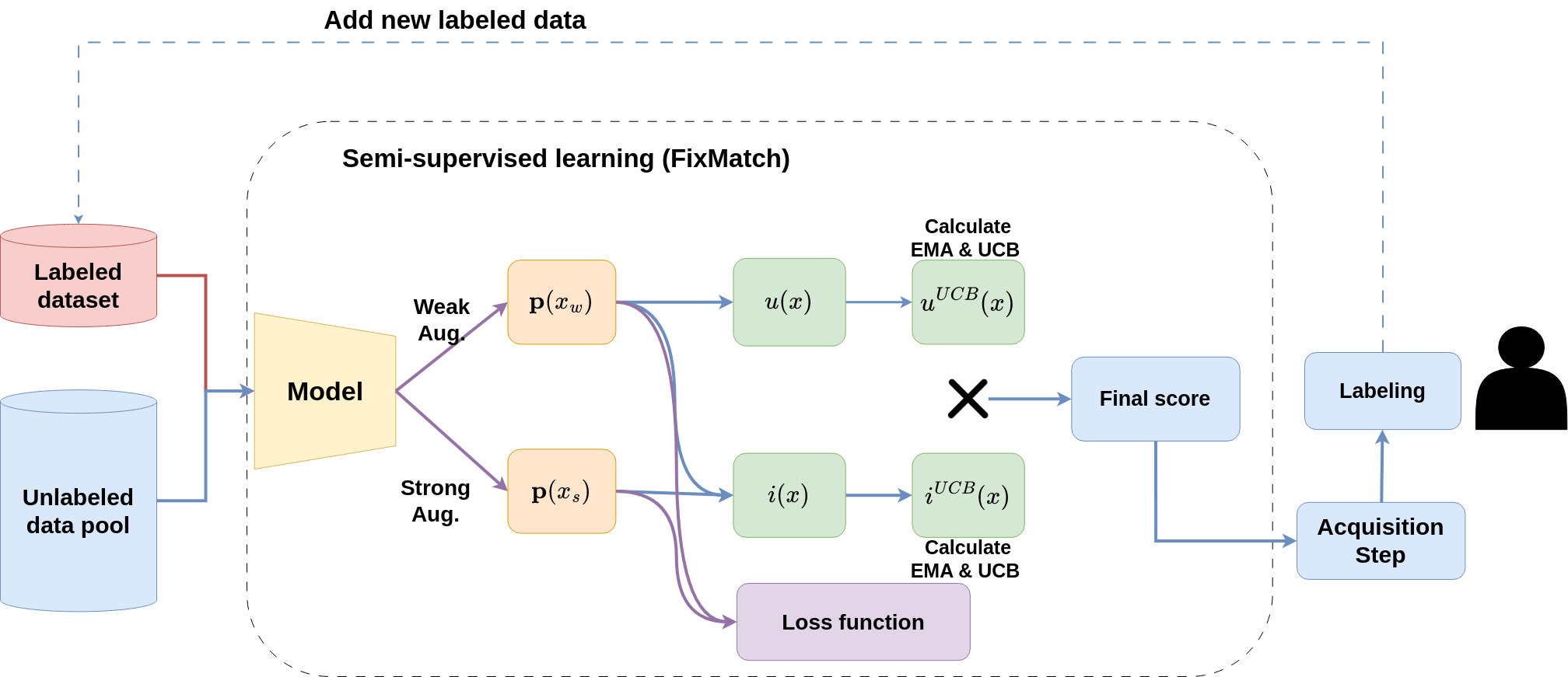}
\end{center}
   \caption{The overview and scoring flow of our proposed method. red line is flow for labeled data, blue line is flow for unlabeled data and purple line is flow for both. Our proposed method calculate uncertainty and data-inconsistency of unlabeled data and utilize the exponential moving average and upper confidence bound to achive final active learning score, which was derived from the probability vector of the unlabeled data pool obtained during the SSL process.
   }
\label{fig:overview}
\end{figure*}
AL is a method of selecting a small number of data samples from a large data pool for labeling and using the newly labeled samples as well as the original labeled samples for further learning. Its acquisition function tries to select the most informative samples. This method can increase the performance of the model more efficiently 
than labeling the entire data pool by iteratively updating the model. In deep learning literature, one of the representative methods of AL is uncertainty-based methods that express the informativeness of data samples as scores~\cite{6889457, Yoo_2019_CVPR, 10.1007/11871842_40, pmlr-v70-gal17a}, while diversity-based methods select data samples that generalize the domain representation of the entire data pool well in the feature embedding space~\cite{10.1007/s11263-014-0781-x, sener2018active}. Recently, research has been conducted on combining these two methods~\cite{10.1007/978-3-030-58517-4_9, Ash2020Deep, Parvaneh_2022_CVPR} and utilizing methods such as adversarial networks~\cite{Kim_2021_CVPR, Sinha_2019_ICCV, pmlr-v108-shui20a}.

Unlike AL, SSL directly utilizes large unlabeled data without an additional labeling process to improve the accuracy of a model with limited labeled data. Consistency regularization methods~\cite{tarvainen2017mean, berthelot2019remixmatch, berthelot2019mixmatch, sohn2020fixmatch, miyato2018virtual, xie2020unsupervised} are the most popular methods in SSL, that enforce the outputs of pairs of differently augmented unlabeled (and labeled) samples to be similar. FixMatch~\cite{sohn2020fixmatch} is the most representative method of this kind, which uses weak and strong augmentation pairs and pseudo-labeling to generate supervisory signals for unlabeled data.

Recently, to present a more powerful and efficient learning methodology by combining AL and SSL, active semi-supervised learning (ASSL)~\cite{gao2020consistency, Elezi_2022_CVPR, kong2022a, https://doi.org/10.48550/arxiv.1912.00594} has been researched. These methods use the data pool of AL as the unlabeled data for SSL and have shown outstanding performance in object classification and detection tasks. However, as Oliver \etal\cite{oliver2018realistic} pointed out, training an SSL algorithm typically requires a very large number of training steps ($\sim 500k$), and if SSL is simply used in each iteration of AL, too much computational resource is required, which can prohibit ASSL from practical usage. However, since SSL can still yield better results than supervised learning using a small number of labeled data, 
if utilized appropriately, it can be combined with AL resulting in an efficient ASSL scheme. Therefore, we tried to combine AL's acquisition step with a smaller number of training steps than typical SSL training to achieve more accurate and faster performance than traditional AL or SSL.

However, there are two main challenges in directly combining AL and SSL. First, calculating the uncertainty of unlabeled samples is quite difficult. In the training of SSL, unlabeled data cause more dynamic model updates compared to supervised learning since the target for a sample is not fixed during training. 
This can lead to catastrophic forgetting~\cite{kirkpatrick2017overcoming} to unlabeled data.
In other words, 
at the moment a specific unlabeled mini-batch is used for learning, the prediction of other unlabeled samples is affected, and this means that the uncertainty score of the unlabeled samples changes according to the acquisition time of ASSL.
This phenomenon also occurs even after a significant number of training steps of ASSL, making it difficult for ASSL to view the observed uncertainty value as truly informativeness of the data. For example, as can be seen in Fig.~\ref{fig:2}, uncertainty-based methods show poorer performance than random sampling in ASSL.
We name this problem as `\textit{temporal-instability}' which highlights that the prediction of unlabeled data suffers from time-dependent changes. To tackle this issue, we calculate the uncertainty for each unlabeled mini-batch and apply a \text{moving average} to minimize the effect of temporal-instability in data scoring, as shown in Fig.~\ref{fig:overview}. Furthermore, we borrow the concept of \textit{upper confidence bound} (UCB) from reinforcement learning to acquire more informative samples based on the variance of uncertainty score.

Second, SSL uses two types of loss functions, supervised loss and unsupervised loss, for model updates through backpropagation. However, uncertainty only focuses on informative samples for the supervised loss, and does not consider the impact of unlabeled data on the unsupervised loss. Therefore, we measure the \textit{data-inconsistency} of the weak and strong augmentation pairs of unlabeled samples, similarly as in \cite{Elezi_2022_CVPR, gao2020consistency}, to analyze the impact of unlabeled data on unsupervised loss. Through this, we identify data samples that give noisy supervisory signals during the consistency regularization process and incorporate this into the acquisition function to improve not only the performance of supervised learning but also that of unsupervised learning.
Our main contributions can be summarized as follows:
\begin{itemize}[leftmargin=*]
    \item We propose a practical ASSL workflow to reduce the training cost, which speeds up training more than 6 times  compared to full SSL training that takes around $500k$ steps.
    \item We observed the problem of instability in measuring the uncertainty of a sample in ASSL caused by temporal-instability and resolved it by exploiting the exponential moving average (EMA) and its upper confidence bound (UCB) to acquire the informative samples in ASSL.
    \item We utilize data-inconsistency of weak \& strong augmentation pair and detect unlabeled samples that make noisy supervisory signal in ASSL. 
    \item We propose a new acquisition function in ASSL by comprehensively considering samples' uncertainty in prediction and data-inconsistency and the effectiveness of the proposed method is demonstrated through thorough experiments.
\end{itemize}

\section{Related works}
\subsection{Active learning}

Active learning is traditionally divided into two categories: uncertainty-based and diversity-based approaches. Uncertainty-based methods involve measuring the uncertainty of the unlabeled data pool and selecting the $N$ samples with the highest uncertainty. Uncertainty can be determined by computing the entropy of the probability of a data sample \cite{6889457}, the margin of the first and second predicted probability \cite{10.1007/11871842_40}, or by using Bayesian deep neural networks with Monte Carlo (MC) dropout \cite{pmlr-v70-gal17a}. More recent studies, such as LLAL \cite{Yoo_2019_CVPR}, predict the loss of a data sample. Although uncertainty-based methods are relatively simple and have demonstrated excellent results, they tend to repeatedly select data samples with specific features, leading to poor generalization performance on the actual data pool since they do not consider the distribution of the data pool.

Diversity-based methods select data samples that represent the overall distribution of the data pool. For instance, studies have investigated clustering the unlabeled data pool to increase diversity \cite{yang2015multi} and creating a CoreSet  of the data pool \cite{sener2017active}. These methods do not suffer from the limitations of uncertainty-based approaches, but they are influenced by the density of the data pool and have limited success in improving the decision boundary. Furthermore, since they require comparing the distance between feature embeddings, they have the disadvantage of relatively high computation.

Recently, there has been an emergence of hybrid methods that combine both uncertainty-based and diversity-based approaches. CDAL \cite{10.1007/978-3-030-58517-4_9}, for example, introduces a weighted probability distribution to define the context of data samples that applies the uncertainty of data samples and aims to maximize diversity between the contexts of data samples. BADGE \cite{Ash2020Deep}, on the other hand, proposes gradient embedding by combining the gradients of data samples with feature embeddings and uses a strategy of clustering and selecting center points. ALFA-Mix \cite{Parvaneh_2022_CVPR} measures the sensitivity of a data sample by interpolating labeled and unlabeled sample feature embeddings and selecting center points of sensitive candidates.

\subsection{Semi-supervised learning}

Consistency regularization \cite{sajjadi2016regularization} is one of the most widely used methods for semi-supervised learning. Recent approaches of this kind have focused on regularizing consistency between weak and strong augmentations of unlabeled data \cite{xie2020unsupervised, berthelot2019remixmatch}, and FixMatch \cite{sohn2020fixmatch} have combined consistency regularization with pseudo-labeling \cite{lee2013pseudo}. It serves as a base framework for various methods. For example, Semi-ViT \cite{cai2022semi} utilizes the ViT \cite{dosovitskiy2021an} model with an EMA framework \cite{cai2021exponential} and pseudo-Mixup based on FixMatch. Other methods, such as FlexMatch \cite{zhang2021flexmatch}  and FreeMatch \cite{wang2023freematch}, use curriculum-pseudo-labeling and self-adaptive thresholding, respectively, to obtain a pseudo-labeling with adaptive threshold. Lastly, CR \cite{Lee_2022_CVPR} proposes contrastive regularization with contrastive learning \cite{khosla2020supervised} of the class cluster, which involves both confident and unconfident pseudo-labels.

\subsection{Active semi-supervised learning}
Active learning (AL) and semi-supervised learning (SSL) both aim to leverage the unlabeled data pool to achieve efficient learning. Therefore, recent research has sought to combine the two algorithms. Song \etal\cite{https://doi.org/10.48550/arxiv.1912.00594} propose combining MixMatch \cite{berthelot2019mixmatch} and margin \cite{10.1007/11871842_40}, while Kong \etal\cite{kong2022a} proposed Neural Pre-Conditioning with the gradient vectors of labeled and unlabeled samples, based on the FixMatch-DARP \cite{kim2020distribution} approach. 
According to Gao \etal\cite{gao2020consistency}, in ASSL, assigning labels to highly inconsistent data samples based on data-inconsistency score is a more effective way to minimize unsupervised loss than acquisition based on uncertainty alone.
Elezi \etal\cite{Elezi_2022_CVPR} demonstrate ASSL in an object detection task. Their method is based on CSD \cite{jeong2019consistency} and uses both uncertainty and data-inconsistency to identify over-confident samples that are mispredicted. This approach produces better acquisition results than using only uncertainty and also uses pseudo-labels to low-score data samples as newly labeled data.

\section{Methods}
\subsection{Problem setting}
In this part, we define the ASSL problem for multiclass classification. Given the labeled dataset $L_n$ and a large unlabeled data pool $U_n$ at round $n$ of active learning, we use both $L_n$ and $U_n$ as labeled and unlabeled data for SSL training. After SSL training, we apply an acquisition function to select $K$ samples $\hat{L}_n$ from $U_n$ for labeling. The labeled dataset at round $n+1$ is then updated as $L_{n+1}=L_n \cup \hat{L}_n$. Likewise, the unlabeled data pool is updated as $U_{n+1} = U_n \setminus \hat{L}_n$.

Let $C=\left\{1,\cdots,k\right\}$, be the set of classes, and let $x$ be an input image. For $y \in C$, we represent the final softmax output of the neural network $f$ for $x$ as $p(y|x;\theta)= \textsc{softmax}(f(x;\theta))$. For simplicity, we denote the vector $[p(y=1|x;\theta), \cdots, p(y=k|x;\theta)]^T$  as $\mathbf{p}(x)$. The predicted label for $x$ is denoted as $\hat{y} = \arg\max(\mathbf{p}(x))$, and the corresponding one-hot vector is written as $\mathbf{1}_{\hat{y}} \in \mathbb{R}^k$.

\subsection{Uncertainty score}
As mentioned in previous works \cite{Ash2020Deep, kong2022a}, in deep neural networks using stochastic gradient descent (SGD), larger gradient magnitudes can result in larger updates to the network parameters. Following \cite{paul2021deep}, which approximated the norm of the gradient with the Error L2-Norm (EL2N), which is the L2-norm of the difference between $\mathbf{p}(x)$ and the corresponding ground truth one-hot vector, we also approximate the gradient for unlabeled data using the predicted label $\hat{y}$.
More specifically, we define the uncertainty score $u(x)$ for input image $x$ as the pseudo-EL2N score between the one-hot vector of $\hat{y}$ and $\mathbf{p}(x)$ as
\begin{equation} \label{eq:1}
u(x) = \left\| \mathbf{p}(x)-\mathbf{1}_{\hat{y}}\right\|_2.
\end{equation}
 \begin{figure}[t]
\begin{center}
 \begin{subfigure}{0.45\linewidth}
 \begin{center}
 \includegraphics[width=1.0\linewidth]{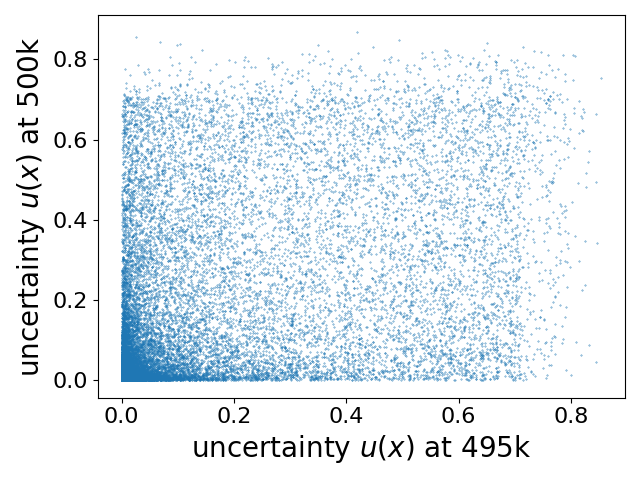}
\caption{} \label{fig:3_a}
\end{center}
\end{subfigure}
 \begin{subfigure}{0.45\linewidth}
 \begin{center}
 \includegraphics[width=1.0\linewidth]{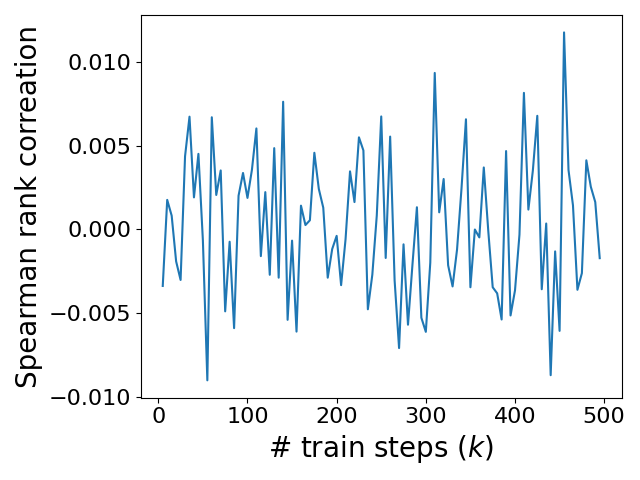}
\caption{} \label{fig:3_b}
\end{center}
\end{subfigure}
\end{center}
   \caption{(a) The distribution of uncertainty(pseudo-EL2N) within 495k training step and 500k training step from unlabeled data pool. (b) Spreaman rank correlation coefficient between the uncertainty scores obtained by current and privious results of every $5k$ training step, there is almost no correlation of two training step.}
\label{fig:3}
\end{figure}
\subsection{Temporal-instability}
 \begin{figure}[t]
\begin{center}
\includegraphics[width=0.8\linewidth]{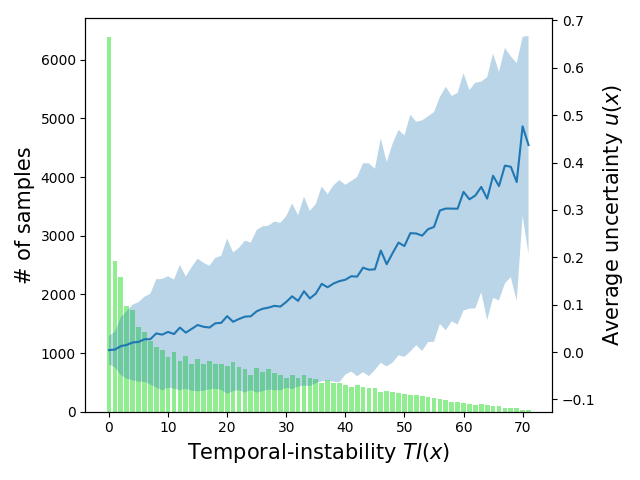}
\end{center}
   \caption{Mean and variance of uncertainty score according to temporal-instability. The green bars indicate the number of data for each temporal-instability value.}
\label{fig:4}
\end{figure}
As shown in Fig.~\ref{fig:2}, when using SSL in the framework of uncertainty-based AL approaches (entropy \cite{6889457}, pseudo-EL2N), the acquisition quality is significantly reduced compared to random sampling, unlike AL with supervised learning. To analyze this problem, we observed the prediction changes in the unlabeled data during SSL training using 100 randomly labeled data. We  calculated the uncertainty $u(x)$ in Eq.(\ref{eq:1}) 
for unlabeled data $U$ every 5k training step and observed how consistent the uncertainty was over time.
Surprisingly, as shown in Fig.~\ref{fig:3}, the uncertainties of consecutive time-stamps showed almost no correlation. 
This phenomenon was consistent throughout all stages of SSL training including even after the model performance sufficiently converged. In other words, the weight parameters of the model continued to be dynamically updated during training SSL. We called this phenomenon as \textit{temporal-instability} in SSL.

In 
\cite{toneva2018empirical}, forgetting events were proposed to observe the changes in prediction and catastrophic forgetting \cite{kirkpatrick2017overcoming} during model updates in supervised learning. Inspired by this, we quantitatively measured the temporal-instability of unlabeled data by measuring the number of times the predicted label $\hat{y}_t(x)$ observed at time $t$ was different from $\hat{y}_{t-1}(x)$ observed at time $t-1$, i.e,
\begin{equation}
TI_T(x) = \sum_{t=1}^T \mathbbm{1}(\hat{y}_{t}(x) \neq \hat{y}_{t-1}(x)),
\end{equation}
where $\mathbbm{1}$ denotes the indicator function.
Fig.~\ref{fig:4} shows the strong positive corelation between the uncertainty score $u(x)$ and temporal instability $TI(x)$.
In the figure, the green bar at $n$ denotes the number of samples with $TI(x) = n$, and the blue line is the average $u(x)$ of those samples accompanied by the corresponding standard deviation.
The strong correlation indicates that samples with high-uncertainty are set on high values of $TI(x)$. And except for small $TI(x)$,  data samples are likely to suffer from fluctuations in predictions and thus in its pseudo-labels. Therefore, unlike conventional AL that does not incorporate SSL, the acquisition quality of uncertainty-based methods has decreased in ASSL because we cannot accurately estimate the informativeness of a data sample based on its uncertainty at a specific time.

\subsection{EMA and Upper Confidence Bound}

To measure the uncertainty more accurately, methods such as Gao \etal\cite{gao2020consistency}, BALD \cite{pmlr-v70-gal17a} and ensemble-based AL \cite{beluch2018power} average the uncertainty of multiple inference results. Also, \cite{raju2021accelerating} proposed using exponential moving averages (EMA) and demonstrated the use of the Upper Confidence Bound (UCB) \cite{Sutton1998}, a reinforcement learning technique, to alleviate temporal-instability and maximize acquisition effectiveness with labeled data in a scenario of dynamic model update.   
Because we can already obtain multiple predictions at different time steps for unlabeled data during the SSL training, 
we also measured the uncertainty of unlabeled mini-batches at each training step and calculated the UCB through the EMA method.
The uncertainty of the unlabeled sample $x$ at time $t$ was measured using the weakly augmented image $\mathbf{p}(x_{w})$, and the equation for EMA and UCB of the uncertainty are as follows:
\begin{equation} 
\begin{split}
&\bar{u}_{t}(x) =  \alpha \cdot u_t(x_{w}) + (1-\alpha)\bar{u}_{t-1}\\
& \bar{v}^{u}_{t}(x) =  \alpha \cdot (u_t(x_{w}) - \bar{u}_{t})^{2} + (1-\alpha)\bar{v}^{u}_{t-1} \\
&   u^{UCB}_{t}(x) =  \bar{u}_{t}(x) + c\cdot \sqrt{\bar{v}^{u}_{t}(x)}
   \end{split}\label{eq:2}
\end{equation}
 where $\bar{u}_{t}(x)$ and $\bar{v}_{t}(x)$ are EMA and exponential moving variance of uncertainty at time $t$, which are initialized as $\bar{u}_{0}(x)=0$ and $\bar{v}_{0}(x)=0$. Also $\alpha$ and $c$ are EMA rate and confidence value of UCB. In addition, to adopt EMA and UCB functions to measure uncertainty, we no longer have to infer unlabeled data in the acquisition step and obtain the uncertainty score right after the end of the SSL training step.

\subsection{Data-inconsistency}
In ASSL, choosing data samples with a higher UCB value of uncertainty leads to the selection of samples that can generate a higher supervised loss in the next round. However, in FixMatch, acquisition functions that rely solely on supervised loss have restrictions due to the presence of unsupervised loss from consistency regularization. As a result, we explored the influence of unlabeled data on unsupervised loss.
To measure data-inconsistency, Elezi \etal\cite{Elezi_2022_CVPR} used the Kullback-Leibler (KL) divergence between the original unlabeled image $\mathbf{p}(x)$ and its flipped version $\mathbf{p}(x_{f})$
FixMatch-based methods \cite{sohn2020fixmatch, cai2022semi, zhang2021flexmatch, wang2023freematch, Lee_2022_CVPR, kim2020distribution} use random weak and strong augmentation pairs \cite{xie2020unsupervised, berthelot2019remixmatch} in the consistency regularization process, making it difficult to precisely identify inconsistencies during the acquisition step. Therefore, like uncertainty $u(x)$, we measured data inconsistency for random weak and strong augmented images $x_{w}$ and $x_{s}$ of unlabeled mini-batches during the training step using the KL-divergence as shown in Eq.~(\ref{eq:4}) and applied EMA as in Eq.~(\ref{eq:2}):
\begin{equation} \label{eq:4}
\begin{split}
    i(x) =& \frac{KL\left ( p\left ( x_{w} \right ), p\left ( x_{s} \right ) \right )+ KL\left ( p\left ( x_{s} \right ), p\left ( x_{w} \right ) \right )}{2} \\
   & \bar{i}_t(x) = \alpha \cdot i_t(x) + (1-\alpha) \cdot \bar{i}_{t-1}(x).
   \end{split}
\end{equation}
\begin{table}[]
\begin{center}
\resizebox{0.9\columnwidth}{!}{%

\begin{tabular}{|c|ccc|ccc|}
\hline
\multirow{2}{*}{\begin{tabular}[c]{@{}c@{}}Sorted\\ by\end{tabular}} & \multicolumn{3}{c|}{Data-inconsistency}                                               & \multicolumn{3}{c|}{Uncertainty}                                                      \\
 & Top 1\% & Top 5\%   & Top 10\%   & Top 1\%  & Top 5\%  & Top 10\% \\ \hline \hline
ratio  & \multicolumn{1}{r}{54.93\%} & \multicolumn{1}{r}{47.46\%} & \multicolumn{1}{r|}{44.20\%} & \multicolumn{1}{r}{15.67\%} & \multicolumn{1}{r}{24.87\%} & \multicolumn{1}{r|}{28.14\%} \\ \hline
\end{tabular}%

}
\end{center}
  \caption{Average ratio of pseudo-labeled samples among samples with top $(1/5/10)\%$ data-inconsistency and uncertainty in fully trained SSL (=$500k$ training step)} 
\label{table:2}
\end{table}

Gao \etal\cite{gao2020consistency} pointed out that samples with high data-inconsistency are overly confident but the unsupervised loss based on these samples are hard to minimize. To measure this, similar to $TI(x)$, for each sample, we counted the number of times  $\max(\mathbf{p}(x)) > \tau (= 0.95)$ at every 5k training step during SSL training. We called these samples `pseudo-labeled', which have high confidence, and thus participate in the unsupervised loss. Table~\ref{table:2} reveals that the proportion of peudo-labeled data with high data-inconsistency is approximately twice as high as that with high uncertainty. Interestingly, these highly data-inconsistent samples are utilized more frequently during training, despite not benefiting from consistency regularization. Consequently, such samples could introduce label noise to the unsupervised loss function.

\begin{figure*}[ht!]
\begin{center}
 \begin{subfigure}{0.3\linewidth}
 \begin{center}
 \includegraphics[width=1.0\linewidth]{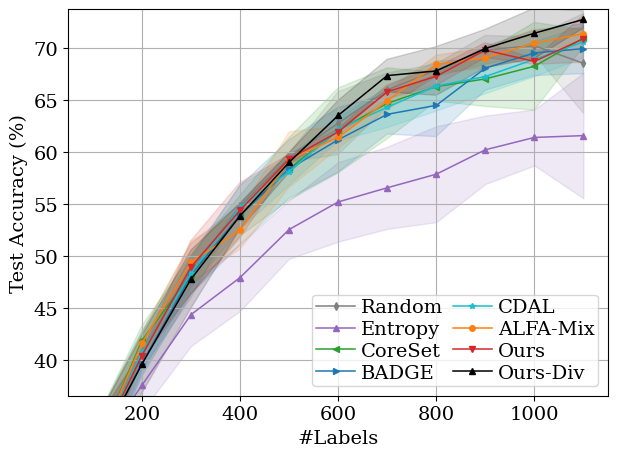}
\caption{CIFAR10, ResNet18, ConInit} \label{fig:res_acc_a}
\end{center}
\end{subfigure}
 \begin{subfigure}{0.3\linewidth}
 \begin{center}
 \includegraphics[width=1.0\linewidth]{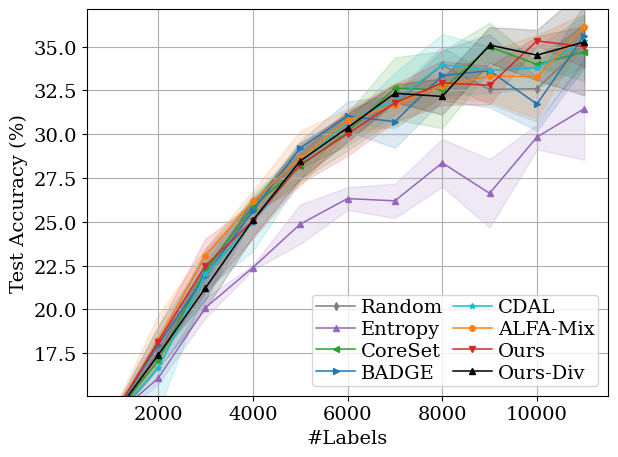}
\caption{CIFAR100, ResNet18, RandInit} \label{fig:res_acc_b}
\end{center}
\end{subfigure}
\begin{subfigure}{0.3\linewidth}
 \begin{center}
 \includegraphics[width=1.0\linewidth]{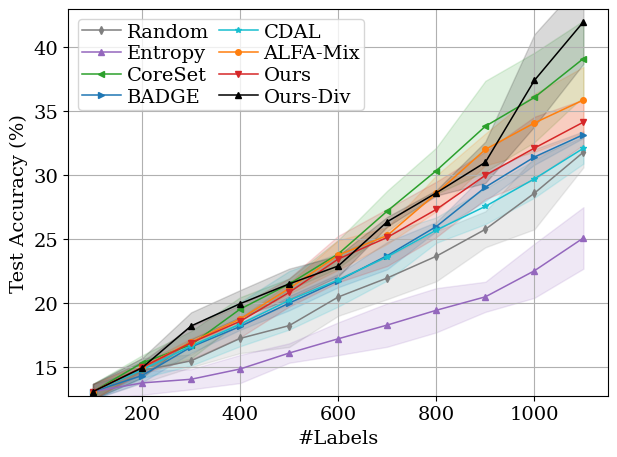}
\caption{SVHN, ViT-small, RandInit} \label{fig:res_acc_c}
\end{center}
\end{subfigure}
\end{center}
   \caption{Comparison accuracy results with various setting.}
\label{fig:res_acc}
\end{figure*}

As same as uncertainty, we utilize UCB as described in Eq.~(\ref{eq:2}) to determine the final data inconsistency score as shown in Eq.~(\ref{eq:i_ucb}). 
\begin{equation} 
\begin{split}
& \bar{v}^{i}_{t}(x) =  \alpha \cdot (i_t(x) - \bar{i}_{t})^{2} + (1-\alpha)\bar{v}^{i}_{t-1} \\
&   i^{UCB}_{t}(x) =  \bar{i}_{t}(x) + c\cdot \sqrt{\bar{v}^{i}_{t}(x)}.
   \end{split}\label{eq:i_ucb}
\end{equation}
Consequently, we define the ultimate AL score by considering both uncertainty and inconsistency in the unlabeled data pool, as described in Eq.~(\ref{eq:5}). 
\begin{equation} \label{eq:5}
   Score(x) =  u^{UCB}(x) \times i^{UCB}(x).
\end{equation}

\section{Experiments}
To evaluate our ASSL method, we used CIFAR-10~\cite{krizhevsky2009learning}, CIFAR-100~\cite{krizhevsky2009learning}, SVHN~\cite{netzer2011reading}, and MiniImageNet~\cite{ravi2017optimization}. For CIFAR-10 and SVHN, we randomly selected 100 images for an initial labeled dataset and acquire 100 additional images for labeling in each round. For CIFAR-100 and MiniImageNet, we started with 1,000 random initial labeled datasets and labeled 1,000 images in each round.

All experiments were conducted with 10 rounds of active learning cycles, and we tested two methods for initial weight condition: random initial weight (RandInit) and the weight from the previous round (ConInit). For RandInit, we used the same random initial weights in all rounds and the initial model. We used the ResNet-18~\cite{he2016deep} and ViT-small~\cite{dosovitskiy2021an} models with the same layer configuration as the model used in ALFA-Mix~\cite{Parvaneh_2022_CVPR}.  We used 1,024 steps $\times$ 5 epochs for training in each round to achieve efficient SSL training, which is approximately $1\%$ of the training step ($500k$) proposed by Oliver \etal\cite{oliver2018realistic}. We set the learning rate as $lr=0.03$ for ResNet-18 and $lr=0.01$ for ViT-small. We did not use learning rate scheduler because of smaller training steps. The other hyper-parameters used in SSL were the same as those in FixMatch~\cite{sohn2020fixmatch}, and we used RandAugment~\cite{cubuk2020randaugment} for augmenting unlabeled data.

We implement our ASSL framework based on ALFA-Mix\footnote {\url{https://github.com/AminParvaneh/alpha_mix_active_learning}} and pytorch implementation of FixMatch\footnote{\url{https://github.com/kekmodel/FixMatch-pytorch}}. For the hyper-parameters of the acquisition function, we set the EMA rate $\alpha$ to 0.8, the confidence value of uncertainty $c_{u}$ to $0.5$, and the confidence value of data-inconsistency $c_{i}$ to $2.0$. We also tried to apply diversity-measure to our proposed method by simply multiplying the score value of each unlabeled sample with its feature embedding vector, and using K-means++~\cite{10.5555/1283383.1283494} for clustering. We denote the proposed method as `Ours', and the method with diversity as `Ours-Div'. The results of all experiments are reported as the average of three runs with different seeds.
\begin{figure}[t]
\begin{center}
\includegraphics[width=0.8\linewidth]{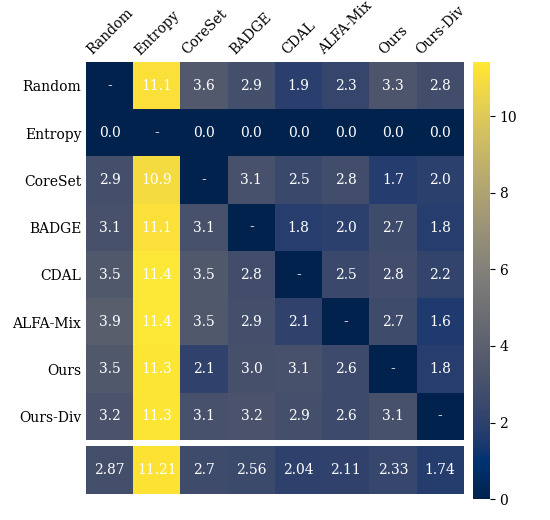}
\end{center}
   \caption{Pairwise comparison matrix of SOTA methods. The overall performance(lower is better) is shown at the bottom row and maximum value is 12.   
}
\label{fig:res}
\end{figure}
\begin{table*}[]
\begin{center}
\resizebox{0.8\linewidth}{!}{%

\begin{tabular}{|l|c|ccc|rrrr|}
\hline
\multicolumn{1}{|c|}{}                          & \multicolumn{4}{c|}{Accuracy (\%)}                                                                                          & \multicolumn{4}{c|}{Time (s)}                                                                                                   \\ \cline{2-9} 
\multicolumn{1}{|c|}{\multirow{-2}{*}{Methods}} & \multicolumn{1}{c|}{RandInit(AL)}   & \multicolumn{1}{c}{RandInit} & \multicolumn{1}{c}{ConInit} & \multicolumn{1}{c|}{Total} & \multicolumn{1}{c}{CIFAR10} & \multicolumn{1}{c}{SVHN}       & \multicolumn{1}{c}{CIFAR100} & \multicolumn{1}{c|}{MiniImageNet} \\ \hline \hline
Random                                          & {27.99$\pm0.84$} & 37.36$\pm1.32$              & 42.49$\pm0.90$             & 39.93$\pm1.11$             & -                           & -                              & -                            & -                                 \\
Entropy                                         & {27.33$\pm1.11$} & 30.74$\pm1.71$              & 38.44$\pm1.42$             & 34.59$\pm1.56$             & 25.50                       & 30.38                          & 26.08                        & 51.34                             \\
CoreSet                                         & {27.35$\pm1.08$} & 36.58$\pm1.76$              & 42.59$\pm1.28$             & 39.59$\pm1.52$             & 37.36                       & 43.63                          & 128.13                       & 209.38                            \\
BADGE                                           & {28.28$\pm1.02$} & 37.15$\pm1.43$              & 42.01$\pm1.59$             & 39.58$\pm1.51$             & 204.65                      & 245.87                         & \textgreater 1 hour          & \textgreater 1 hour               \\
CDAL                                            & {28.45$\pm0.97$} & 37.02$\pm1.39$              & 42.62$\pm1.07$             & 39.82$\pm1.23$             & 28.39                       & 34.04                          & 64.06                        & 135.39                            \\
ALFA-Mix                                        & {28.34$\pm1.04$} & 37.53$\pm1.43$              & 42.47$\pm0.90$             & 40.00$\pm1.17$             & 83.17                       & 104.76 & 490.95                       & 1013.44                           \\ \hline
Ours                                            & {-}               & 37.07$\pm1.44$              & \textbf{42.80}$\pm0.81$    & 39.93$\pm1.13$             & 0.05                        & 0.06                           & 0.03                         & 0.05                              \\
Ours-Div                                        & {-}               & \textbf{37.68}$\pm1.49$     & 42.62$\pm1.04$             & \textbf{40.15}$\pm1.27$    & 36.05                       & 43.39                          & 129.14                       & 209.10                            \\ \hline
\end{tabular}%
}
\end{center}
\caption{Average accuracy in terms of initial weights condition and aqcusition times}
\label{table:3}
\end{table*}
\subsection{Results}
To evaluate our proposed method, we compared our methods with random sampling and state-of-the-art AL methods such as entropy~\cite{6889457}, BADGE~\cite{Ash2020Deep}, CDAL~\cite{10.1007/978-3-030-58517-4_9}, CoreSet~\cite{sener2017active}, and ALFA-Mix~\cite{Parvaneh_2022_CVPR}. We evaluate them by using Resnet-18 on four aforementioned datasets. Furthermore, to test if our method works on different architectures, we compare the result of ViT-small on CIFAR10 and SVHN. To comprehensively analyze the experimental results, we compute the pairwise comparison matrix~\cite{Ash2020Deep} in Fig.~\ref{fig:res} that each elements $e_{i, j}$ indicates that the number of i-th method outperformed j-th and its symmetric elements $e_{j, i}$ is the opposite case. the last row at bottom is average score by column that lower score means better.
As shown in Fig.~\ref{fig:res}, `Ours-Div' achieved the best performance in the combination of four datasets and two models. Also `Ours' shows better results than random sampling in ASSL and outperformed CoreSet and BADGE that rely on diversity. For more details, as shown in Fig.~\ref{fig:res_acc}, 

'Ours' shows comparable performance to state-of-the-art (SOTA) methods, and in some cases, even shows better performance. 

The left part of Table \ref{table:3} is the average  test accuracy of all rounds and datasets by two initial weight parameter conditions. `Our-Div' shows the best average accuracy with Randinit and Total, and `Ours' shows the best average accuracy with ConInit.
 Also, we tested conventional AL frameworks to compare the accuracies between AL and ASSL. We train the AL framework with the same hyperparameter setting that was used by ALFA-Mix~\cite{Parvaneh_2022_CVPR} and average test accuracies as done in ASSL with RandInit condition. Our ASSL framework shows average  accuracy improvements of 8.11\%p compared to all state-of-the-art AL methods.

 Moreover, diversity-based methods, such as CoreSet~\cite{sener2017active}, show relatively low sensitivity to temporal-instability in the setting of ASSL. This means that even if $\mathbf{p}(x)$ of data samples is temporally unstable, the representation of the entire domain is relatively preserved. In other words, $\mathbf{p}(x)$ is expected to change within a specific range without random direction and instability. So adapting EMA and UCB helps to find the maximum expected value in the unstable area.
 
\subsection{Training efficiency}
 \begin{figure}[t]
\begin{center}
\includegraphics[width=0.8\linewidth]{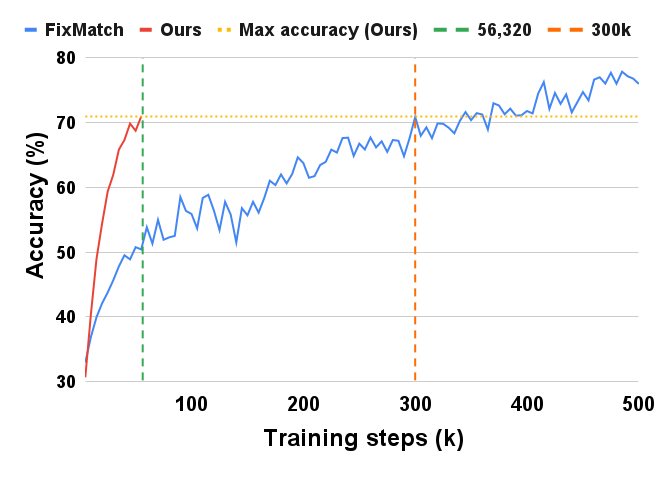}
\end{center}
   \caption{Compare accuracies of FixMatch and Ours according to training steps}
\label{fig:ssl}
\end{figure}

In terms of training time, the proposed method was compared to FixMatch. Up to 10 rounds, the total number of training steps was 56,320, which represents about 11\% of the 500$k$ training steps. Fig.~\ref{fig:ssl} compares the test accuracy of FixMatch and Ours during training. Using CIFAR-10 on the Resnet-18 model, FixMatch reached 70.82\%p at 300$k$ training steps which is the closest to 70.90\%p obtained by Ours(ConInit) at around 56k training steps. In this case,  our proposed method is about 5.3 times faster than SSL. In our Nvidia Titan X environment, it took approximately 20 hours for our ASSL to achieve the best accuracy, while SSL required about 4.4 days.

In actual active learning scenarios, data labeling incurs an additional time cost. However, annotating approximately 100 images does not require significant effort. In our experiment, three people spent less than 5 minutes on this task. Therefore, in practice, the small amount of labeling work required by ASSL is a more efficient approach for achieving the target accuracy compared to fully converging the model through SSL.

In terms of acquisition time, as shown in the right section of Table \ref{table:3}, `Ours' calculates the score during the training process, so it can be confirmed that there is almost no time cost in the acquisition step. In addition, similar to CoreSet~\cite{sener2017active}, `Ours-Div' also has a faster acquisition time compared to BADGE~\cite{Ash2020Deep} or ALFA-Mix~\cite{Parvaneh_2022_CVPR} whose time costs depend on the number of classes.

\subsection{Ablation study}
\subsubsection{Uncertainty and its UCB}
\begin{figure}[t]
\begin{center}
 \includegraphics[width=0.8\linewidth]{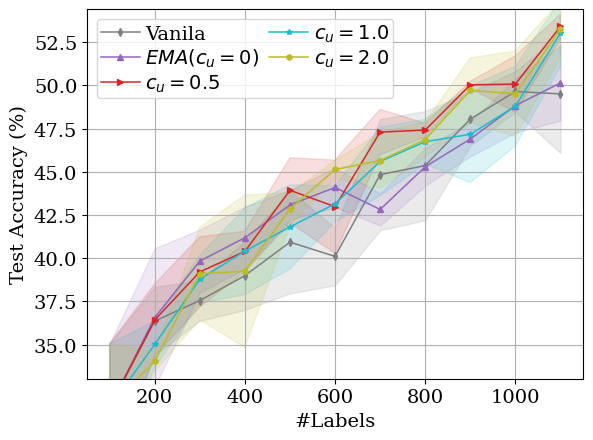}
\end{center}
   \caption{Pseudo-EL2N and its UCB with various confidence value in CIFAR10 and RandInit}
\label{fig:uncer}
\end{figure}

In the ablation study, we analyzed the effects of EMA and UCB of uncertainty (Pseudo-EL2N) with CIFAR-10, as shown in Fig.~\ref{fig:uncer}. We found that using EMA achieves a better average accuracy than vanilla Pseudo-EL2N, and we observed that the accuracy score was higher when additionally using UCB. The average accuracy is the highest when the confidence value of UCB is set to $c_u = 0.5$. Therefore, when the model updates dynamically like in SSL, scoring through moving average is an effective method, and applying additional UCB shows that samples with large supervised loss were selected.

\subsubsection{Data-Inconsistency and its UCB}
\begin{figure}[t]
\begin{center}
\includegraphics[width=0.8\linewidth]{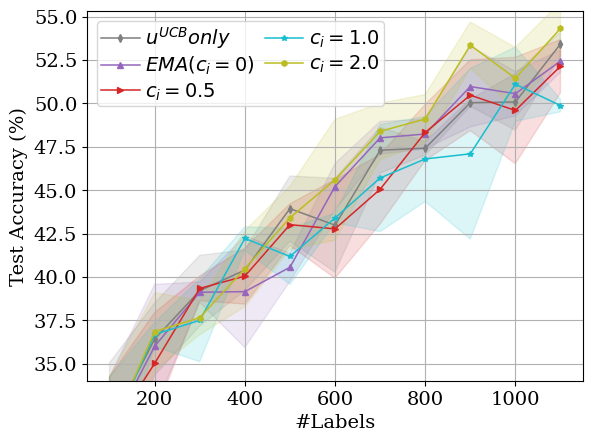}
\end{center}
   \caption{Adapt Data-inconsistency with various confidence value in CIFAR10, RandInit and $c_{u}=0.5$}
\label{fig:inc}
\end{figure}

To analyze the effect of data inconsistency, as shown in Figure \ref{fig:inc}, we found that $c_{i}=2.0$ provided the best performance when combined with $u^{UCB}$. This means that using $i^{UCB}$ with high confidence values is most effective, as data-inconsistency is important to have high values while also having high variance. Reducing the instability of data-inconsistency by UCB improves accuracy than considering only EMA of data-inconsistency.

\section{Discussion}
\textbf{Trade-off of labeling cost and training cost.} In this paper, we conducted experiments using same number of data samples as AL framework for comparison. However, we believe that it is possible to achieve higher accuracy with less labeling cost by leveraging training steps of SSL. On the other hand, simply acquiring more labeled samples would decrease the training cost to achieve the same accuracy. So there would be a trade-off relationship between increasing labeling cost to use fewer training cost. Therefore, adjusting the balance between the two costs according to each task would make our proposed method more efficient.

\textbf{Data-inconsistency.} Through this study, we began to examine how unlabeled data works in SSL. By identifying data inconsistency, we observed unlabeled samples that are not trainable and create noisy supervisory signals. By considering data inconsistency in the acquisition steps, we obtained better results. However, we did not deeply investigate how and to what extent such data samples have a negative impact on SSL. For example, ignoring high data-inconsistency samples during the ASSL and SSL training, or simply removing those samples from unlabeled data pool during the acquisition steps. Additionally, we did not explore whether low data inconsistency data is useful for training. Therefore, we believe that further research and supplementation of these aspects can lead to the development of new methodologies in ASSL and SSL.

\textbf{Apply to other tasks.} As Elezi \etal\cite{Elezi_2022_CVPR} proposed, ASSL is tried to apply to object detection tasks. Also semi supervised learning methods based on consistency regularization has been studied continuously for many computer vision tasks, such as segmentation \cite{Liu_2022_CVPR2} and human pose estimation \cite{https://doi.org/10.48550/arxiv.2203.07837, Xie_2021_ICCV} that suffer high labeling costs than object classification and object detection. We believe that if we apply the proposed ASSL method to the above methods. we can update model fast with more reasonable costs as object classification.
\section{Conclusion}

We present a novel approach to Active Semi-supervised Learning (ASSL) that effectively addresses temporal instability. ASSL faces a challenge in accurately determining the informativeness of unlabeled data using conventional uncertainty-based functions due to temporal instability. To tackle this challenge, we leverage the exponential moving average (EMA) and its upper confidence bound (UCB) of uncertainty to identify truly informative data samples. Additionally, we introduce data-inconsistency to identify samples that are not trainable and incorporate both factors comprehensively in the acquisition function. Our experimental results demonstrate that our method outperforms state-of-the-art active learning methods, with an 8.11\%p improvement over existing AL methods~\cite{6889457, sener2017active, Ash2020Deep, 10.1007/978-3-030-58517-4_9, Parvaneh_2022_CVPR}, while also being approximately 5.3 times more efficient than semi-supervised learning. Going forward, we plan to focus on developing more practical ASSL methods with a smaller number of labeled data and adapting them to other tasks.

{\small
\bibliographystyle{ieee_fullname}
\bibliography{ref}
}

\end{document}